\documentclass[sigconf]{acmart}
\AtBeginDocument{%
  }

\usepackage{xspace}
\usepackage{booktabs}
\usepackage{array}
\usepackage{tcolorbox}
\usepackage{xcolor}
\usepackage{graphicx}

\newcommand{\system}{\textsc{MathVC}\xspace}

\usepackage{soul}

\setcopyright{acmlicensed}
\copyrightyear{2018}
\acmYear{2018}
\acmDOI{XXXXXXX.XXXXXXX}
\acmConference[Conference acronym 'XX]{Make sure to enter the correct
  conference title from your rights confirmation email}{June 03--05,
  2018}{Woodstock, NY}
\acmISBN{978-1-4503-XXXX-X/2018/06}

\newcommand{\takeaway}[1]{
\noindent\rule{\linewidth}{0.05pt}
\par\nobreak\noindent\textbf{Takeaways:}
#1
\vspace{-2mm}
\par\nobreak\noindent
\rule{\linewidth}{0.05pt}
}

\begin{document}

\title{MathVC: An LLM-Simulated Multi-Persona Virtual Classroom for Collaborative Mathematical Problem Solving} 

\author{Murong Yue}
\authornote{Both authors contributed equally to this research.}
\affiliation{%
  \institution{George Mason University}
  \city{Fairfax}
  \country{USA}}
\email{myemail@gmu.edu}

\author{Wenhan Lyu}
\authornotemark[1]
\affiliation{%
  \institution{William \& Mary}
  \city{Williamsburg}
  \country{USA}}
\email{wlyu@wm.edu}

\author{Jennifer Suh}
\affiliation{%
  \institution{George Mason University}
  \city{Fairfax}
  \country{USA}}
\email{jsuh4@gmu.edu}

\author{Yixuan Zhang}
\affiliation{%
  \institution{William \& Mary}
  \city{Williamsburg}
  \country{USA}}
\email{yzhang104@wm.edu}

\author{Ziyu Yao}
\affiliation{%
  \institution{George Mason University}
  \city{Fairfax}
  \country{USA}}
\email{ziyuyao@gmu.edu}


\renewcommand{\shortauthors}{Yue et al.}

\begin{abstract} 
Collaborative problem-solving (CPS) is essential in mathematics education, fostering deeper learning through the exchange of ideas. Yet, classrooms often lack the resources, time, and peer dynamics needed to sustain productive CPS. Recent advancements in Large Language Models (LLMs) offer a promising avenue to enhance CPS in mathematical education. We designed and developed \system, a multi-persona LLM-simulated virtual classroom platform to facilitate CPS in mathematics. \system combines a meta‑planning controller that monitors CPS stages—sense‑making, team organization, planning, execution, validation—and predicts the next speaker, with a persona‑simulation stack that encodes mathematical thinking via a task schema and error‑injected persona schemas seeded from teacher‑specified misconceptions. We evaluated \system with 14 U.S. middle‑schoolers. Students reported constructive interaction and reaching shared solutions, describing gains in engagement, motivation, and confidence through diverse perspectives, immediate scaffolding, and human‑like fallibility. Our findings also provide insights into simulating peers via LLM-based technologies for collaboration to support learning. 
\end{abstract}



\maketitle

\section{Introduction}

Collaborative problem-solving (CPS), i.e, the capacity to effectively engage in a process whereby two or more students attempt to solve a problem by sharing the understanding and effort required to come to a solution, and pooling their knowledge, skills, and efforts to reach that solution~\cite{pisa2018assessment}, is central to mathematics education. Students learn most when they exchange ideas, test strategies, and refine reasoning collaboratively~\cite{polya2014solve,schoenfeld2016learning}. In middle school, CPS is especially important because problems become more open-ended and require multiple perspectives~\cite{gauvain2018collaborative}.  For middle school students, cognitively, they are transitioning from concrete to abstract thinking, capable of handling multi-step and open-ended problems through iterative refinement and critical thinking. Middle school mathematics tasks encourage rich ``math talk,'' including constructing arguments, critiquing reasoning, and questioning peers—core competencies outlined in the Standards for Mathematical Practice
~\cite{klibanoff2006preschool}.  
However, real classrooms often struggle to achieve ideal CPS conditions due to diverse student backgrounds (e.g., varying demographics, math proficiencies, personalities, etc.) and large class sizes. For example, CPS can be dominated by students with high competencies or who are more confident or extroverted in conversations, whereas less capable, less confident, or introverted students do not receive equal opportunities to participate in the discussion and practice their math thinking. On the other hand, 
large class sizes inevitably leave many students on the sidelines or significantly increase the orchestration effort from teachers to monitor each CPS group, making consistent engagement across the class challenging.
As a result of these circumstances, student participation has become imbalanced, leading to differing degrees of effective learning in CPS \cite{Yang2024}.

Previous research has introduced tools aimed at fostering balanced group participation during CPS, supporting less vocal or less confident students through technologies, such as tangible interfaces and AI facilitators~\cite{bachour2010interactive,cheng2022investigation,li2022designing, liu2023learning, siledar2024one}. However, these systems largely optimize the process rather than the people who populate that process, which does not address issues of uneven participation and the lack of effective math learning moments. Instead, modeling student peers and directly creating a well-orchestrated environment at the peer level becomes important, so that students can engage in CPS conversations that naturally elicit clarification, critique, and collaborative reasoning, even when suitable human peers are unavailable~\cite{nelson2013collaborative}.

Large Language Models (LLMs) make such peer simulation feasible, and recent work has explored ``LLM personas'' across domains~\cite{Park2023GenerativeAgents,wang-etal-2023-humanoid, zhou2023characterglm}. These personas are capable of generating believable human-like interactions in simulated environments, such as social communities~\cite{Park2023GenerativeAgents, 10.1145/3613904.3642406, 10.1145/3706598.3714034}, embodied agents~\cite{wang-etal-2023-humanoid, 10.1145/3613904.3642406}, and teaching assistant training~\cite{Markel2023GPTeachIT, 10.1145/3613904.3642377, 10.1145/3613904.3642773}. However, little work has specifically examined how LLM personas can be designed to support (mathematical) CPS scenarios.
Furthermore, prior evaluations of LLM persona (and broadly, LLM agents) have predominantly focused on automated performance benchmarks rather than real-world interactions involving targeted end users~\cite{li2025exploring, chu2025llm}. In parallel, HCI scholars have called for more empirical studies to evaluate how human subjects interact with LLM agents in real-world interactions~\cite{10.1145/3706599.3716299, 10.1145/3711015, 10.1145/3706598.3713726, 10.1145/3706599.3706729}. Such empirical evaluation will allow us to observe engagement patterns, learning gains, and breakdowns that benchmarks cannot reveal. Our work seeks to address both research gaps by \textit{designing} and \textit{empirically evaluating} a multi-persona virtual classroom platform with real students engaged in collaborative mathematical problem-solving.



In this work, we designed and developed \textbf{\system, a Mathematics Virtual Classroom in which a middle school student converses with several LLM-simulated ``virtual peers'' for mathematical CPS}. Unlike prior work, which enhanced collaboration mainly by \emph{assisting} the process,
\system directly \emph{creates a collaborative environment} that allows students to participate in mathematical CPS. The key innovation of \system lies in leveraging multiple virtual peers---each simulated by distinct LLM-driven personas---to engage students in mathematical CPS. 
In \system, one student collaborates with multiple virtual peers, and each virtual peer offers its own solution ideas, questions, and critiques, so the student can negotiate meaning, justify reasoning, and refine answers. \system enables students to practice CPS individually in a self-centered, automatically orchestrated environment. 

We then conducted an evaluation study with 14 middle‑school participants to examine their interaction experiences, perceived authenticity of \system, and their views on the \system's role in supporting collaborative mathematical problem-solving. Our findings reveal that \system enhanced students' engagement, motivation, and mathematical confidence through their collaboration with virtual peers. Specifically, participants appreciated virtual peers' realistic collaborative behaviors, such as human-like fallibility (e.g., making errors and correcting them) and authentic social
``voice'' (e.g., sound and act like middle-schoolers). Additionally, students highlighted the critical role of socio-emotional and motivational dimensions when engaging in collaborative mathematical problem-solving with \system. They valued \system's virtual peers in reducing cognitive burden, diverse perspectives supported by LLM personas, and providing supportive, immediate scaffolding, which collectively enhanced their motivation and persistence in mathematical problem-solving. Our participants also found that interacting with \system enhanced their confidence through supportive scaffolding and gained their metacognitive confidence through articulation and error handling by \system.

\textbf{Contributions:} 
We contribute \textbf{empirical explorations} examining \textbf{multi-persona LLM systems as collaborative peers for mathematical problem-solving} with \textbf{evaluation involving real students}. Specifically, we design and develop \system, a multi-persona, LLM-based virtual classroom tailored to middle-school collaborative mathematical problem-solving. Second, through conducting an empirical evaluation study with middle school learners, we gain insights into their interaction experiences, perceived authenticity, and their attitudes and possible impacts of using \system on their socio-emotional and motivational engagement in CPS in mathematics. Third, our findings provide design implications for building future LLM‑based collaborative learning tools and for leveraging LLM-simulated personas into mathematics education.

\section{Related Work}

\subsection{Collaborative Problem Solving}
Collaborative problem-solving can be defined as the ``\textit{capacity to effectively engage in a process whereby two or more agents attempt to solve a problem by sharing the understanding and effort required to come to a solution, and pooling their knowledge, skills, and efforts to reach that solution}'', according to the Programme for International Student Assessment (PISA)~\cite{pisa2018assessment}. 
The HCI community has explored technologies and design principles for supporting CPS. Prior HCI studies highlight the significance of shared awareness, coordination, and mutual understanding as critical factors influencing effective collaboration~\cite{gutwin2002descriptive,olson2000distance}. Furthermore, research emphasizes how technologies can scaffold diverse social interactions and enable more equitable participation in collaborative processes \cite{borge2018learning,kies1998coordinating,bauer2017collaborative, 10.1145/2818052.2874324}.

In the context of mathematics education, CPS is regarded as a fundamental skill that develops students’ critical thinking abilities~\cite{polya2014solve, schoenfeld2017learning}. In classrooms, students engage in problem-solving through  ``group worthy'' tasks~\cite{cohen2014designing} where tasks are designed to be open-ended. Oftentimes, students with different abilities and perspectives, are expected to contribute to problem-solving ``math talk''~\cite{cohen2014designing}. In 2015, the PISA Collaborative Problem Solving assessments were administered as a large-scale international assessment to evaluate students' competency in collaborative problem solving by requiring them to interact with simulated computer agents designed to represent different team member profiles~\cite{pisa2018assessment}. 
These agents responded to students' inputs following a scripted virtual chat, and the assessment included various types of tasks such as group decision-making, group coordination, and group production to elicit diverse problem-solving behaviors and interactions. The assessment showed promise in employing a technology-enhanced assessment of skills that traditionally required multiple students and extensive monitoring by instructors, providing immediate and focused feedback in a realistic collaborative setting~\cite{csapo2017nature, graesser2017assessment}.

A growing body of work has explored computer-supported collaborative learning systems to support scaffolding collaborative interactions at scale~\cite{liu2020dashboards}.
MathCHOPS~\cite{bergner2023mathchops}, for example, pairs learners and monitors discourse moves to prompt higher-order reasoning during word problem-solving. 
The Graspable Math~\cite{weitnauer2016graspable} shared workspace allows multiple students to manipulate algebraic expressions synchronously and has been linked to richer strategy talk and higher post-test scores. Conceptual frameworks, such as the CoPS model~\cite{fitzsimons2024cops}, outline the specific social, cognitive, and metacognitive moves that predict collaborative success in mathematics.
However, existing systems still depend on real peers; classrooms with diverse student populations (e.g., demographics, varying math proficiencies) continue to struggle with productive group work. For example, digital collaborative math games show benefit only when student peers sustain equitable talk~\cite{10.1145/3341215.3356295, 10.1145/3025453.3025593}, while pairing students for collaborative activities, even with AI support, still adds orchestration overhead for teachers~\cite{10.1145/3544548.3581398}.
Our work explores the potential of LLM technologies in overcoming these challenges. In particular, through \emph{simulating} real peers using LLMs, \system offers the promise of enabling engaging, self-centered, and auto-orchestrated CPS environments for each student learner, which thus encourages effective participation and math learning in CPS.

\subsection{Large Language Models in Mathematics Education}
Prior to the widespread adoption of LLMs, researchers explored methods to design interactive computer agents for tutoring students with crafted templates~\cite{graesser2004autotutor} or integrating latent semantic analysis methods~\cite{graesser2016conversations}.
Advances in LLMs have driven significant shifts in mathematics education across both K-12 and higher education settings~\cite{chu2025llm2edusurvey}.
LLMs assist students by providing detailed explanations \cite{kumar2023math, 10.1145/3613904.3642773, 10.1145/3613904.3642377, 10.1145/3613904.3642647} or asking Socratic questions \cite{al2023socratic, al2024can, 10.1145/3627673.3679881} to facilitate their problem-solving, offering personalized guidance to their learning plan~\cite{scarlatos2025exploring, 10.1145/3706598.3714261}, and identifying and correcting their mistakes~\cite{xu2025ai2errorAnalysis,zhang2025correctness, 10.1145/3613904.3642773}. Systems such as ChatTutor \cite{chen2024empowering} integrate these efforts and aim to offer a comprehensive tutoring process, covering informative instruction, course plan adjustment, quiz offering and evaluation, and more. On the other hand, LLMs also support teachers by designing teaching materials~\cite{koraishi2023teaching,hu2025exploring}, generating exam questions~\cite{lee2024math,nikolovski2025advancing}, and grading assignments~\cite{caraeni2024evaluating,michael2024automatic}. Finally, platforms such as Khanmigo \cite{Khanmigo} provide tools and services facilitating both student learners and teachers. 
Educators and researchers have noted that LLM-based applications can help overcome common challenges in mathematics education, such as limited teacher resources~\cite{darling2013inequality}, by offering readily available feedback and scaffolding.

Besides, recent studies show that LLM-based agents can simulate student personas and offer interactive, personalized support in education, thereby enabling an even more adaptive learning environment. For example, \citet{Markel2023GPTeachIT} simulated students in office hours and applied the system for real-life teaching assistants to receive valuable teaching practices. \citet{xu2024eduagent} collected a dataset of real students' learning behaviors and proposed Eduagent to simulate these behaviors, such that educators can have a more precise modeling and prediction of the students' learning dynamics. To effectively assess the quality of designed questions, \citet{lu2024generative} constructed LLM-simulated students to answer these questions. Finally, in the domain of mathematics education, \citet{10.1145/3706468.3706532} examined the usage of teachable AI agents to facilitate ``learning by teaching'', where students learn by teaching the AI agents. Their work found that the personalities of AI agents have an impact on the students' learning outcomes.

These existing works, however, have predominantly focused on simulating students in teacher-student interactions, yet how to simulate students as \emph{collaborative peers} remains underexplored. As we discuss in Section~\ref{subsec:challenges_simulating}, simulating student peers in a mathematical CPS environment brings unique challenges of aligning the simulated interaction dynamics with the authentic collaboration dynamics among real-life students, which has not been explored in prior works. In addition, unlike teacher-student interactions, CPS often involves more than one student peer, which results in further technical complexity in the simulation modeling.
In this work, we explore approaches to overcome these challenges.
With a carefully designed human study, our work aims to offer new insights into the applications of LLMs to complicated multi-agent settings in mathematics education.

\subsection{Simulating Personas with Large Language Models}

The advancements of LLMs have given rise to the community's increasing interest in using LLMs to simulate personas. 
\citet{chen2024persona} categorized the current research on LLM-powered personas into three main types. 
The first is simulating \textit{demographic personas}, which focuses on simulating groups of people who share common characteristics, such as occupation or ethnic groups. Examples include simulating students, doctors, software engineers, etc~\cite{hong2023metagpt,qian2023chatdev,tang2023medagents}.
The second is to simulate \textit{character personas}, which refers to the simulation of well-known and widely recognized individuals, such as celebrities~\cite{zhou2023characterglm,shao2023characterllm} and virtual characters in books and movies~\cite{wang2025coser}. 
The third type is to build up \textit{individualized personas} for personalized LLM services, such as personal assistants for individuals~\cite{salemi2023lamp} or recommendation systems~\cite{chen2024large}.
Our research aligns with the first category, i.e., simulating demographic personas.

In persona simulation, an LLM is prompted with a specific persona description constructed from the predefined background and persona traits~\cite{shanahan2023role}, enabling it to display human-like behaviors and interactions by drawing on both external prompting instructions and its internal understanding of demographic characteristics~\cite{tseng2024two}.
Prior work has shown that LLM-based demographic persona simulation can be effectively conducted using explicit profiles, episodic memories, and goal structures, enabling believable simulations~\cite{Park2023GenerativeAgents}. For instance, \citet{hong2023metagpt} and \citet{qian2023chatdev} employed LLMs to simulate ``software developers'' collaborating on software projects; \citet{tang2023medagents} simulated AI ``doctors'' engaging in multi-round medical discussions; and \citet{sun2024facilitating} simulated AI ``recruiters'' conducting online hiring processes. 
Complementing existing work, we explore simulating a group of student personas in this work.



Persona simulation is considered faithful when the responses or behaviors produced by the LLM closely mirror those of the corresponding real-world groups~\cite{peng2024quantifying}.
However, a common threat for demographic persona simulation is characteristics drifting: the persona's output deviates from its assigned traits~\cite{fischer2023reflective,wang2024large}, called ``persona misalignment''.
Simulation can become worse when the specified personality or knowledge conflicts with the LLM’s internal tendencies or pretrained knowledge~\cite{kumar2025can}. 
Specifically, since LLMs were typically pre- or post-trained to act as helpful assistants and provide comprehensive and correct answers, instructing them to simulate middle school students with limited knowledge or incorrect solutions presents a significant challenge.
However, there has been little research on LLM-based role-playing involving less knowledgeable virtual classmates who display common misconceptions, thereby providing students with opportunities to identify and correct peer errors—a pedagogical strategy shown to enhance learning in peer tutoring contexts~\cite{csapo2017nature}.
In this work, we tackle the persona misalignment challenge by encoding symbolic persona representations that keep the personas ``in character'' as middle-school peers of varying proficiency and by empirically examining how real students experience these virtual classmates during mathematical CPS tasks.



\section{System Design of \system}
\label{sec:system_design}

\begin{figure*}
    \centering
\includegraphics[width=.9\textwidth]{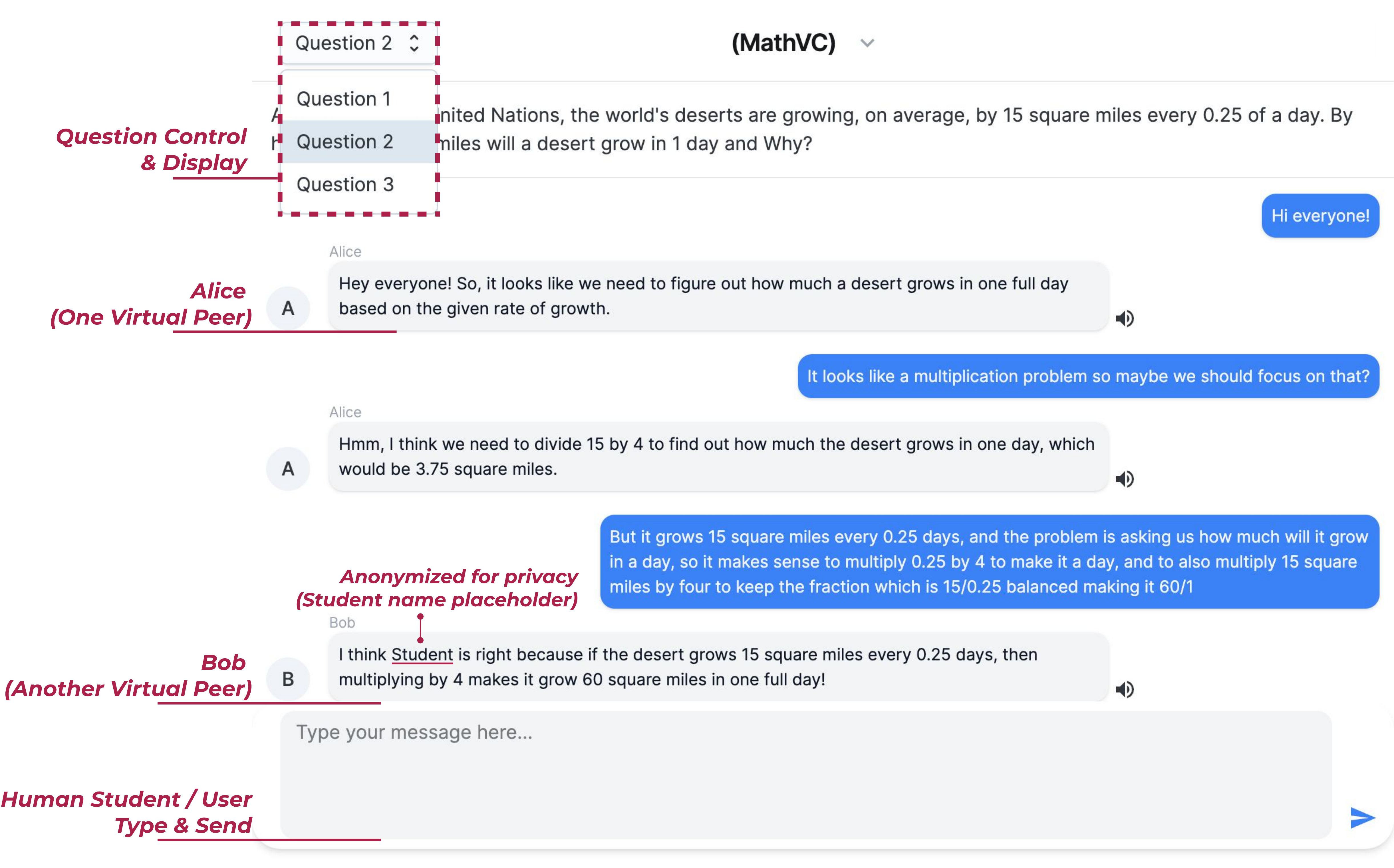}
    \caption{\system user interface has several components: (1) \textit{Question Control and Display}, enabling students to select and navigate among mathematical problems; (2) a \textit{Chat Interface}, where the human student interacts with multiple LLM-simulated virtual peers; and (3) the \textit{User Input Box}, allowing students to send messages and engage collaboratively with virtual peers. In this example session as shown, a middle-school student (anonymized) and two virtual peers (Alice and Bob) collaboratively discuss and solve a mathematical problem.}
    \vspace{-2mm}
    \label{fig: VC}
\end{figure*}


In this section, we present \system, a mathematical virtual classroom we developed where a student engages in collaboratively solving mathematical problems with multiple LLM-simulated student personas.

\noindent\textbf{A Usage Scenario.} 
Imagine that a teacher would like their students to practice more on mathematical problem-solving after class. 
Because of the advanced feature of \system in supporting CPS, the teacher resorts to \system for designing the take-home assignments. 
The teacher can configure appropriate mathematical problems aligned with their curriculum objectives, the common mistakes of human students, and the personal traits of virtual peers, and then grant students access to \system after class.
When a student starts their session, \system automatically sets up the mathematical CPS task and creates virtual peers following the teacher's specification.
During the interaction, the student collaborates with these virtual peers, exchanging solution ideas, asking clarifying questions, and addressing each other's misunderstandings. 
The virtual peers dynamically update their thoughts on the mathematical problem based on the student's inputs and the evolving dialogue, producing responses grounded in their current knowledge and persona traits. The dialogue between the student and virtual peers continues until they resolve the mathematical problem together or when the student decides to move on to the next problem.

To realize this scenario, several challenges must be addressed in the design and simulation of student personas. In the remainder of this section, we outline the key challenges in simulating student behaviors (\autoref{subsec:challenges_simulating}), present our modular architecture for addressing these challenges (\autoref{subsec:modular_arch_design}), and describe system implementation details (\autoref{subsec:implementation}).

\subsection{Challenges of Simulating Middle School Students in CPS}
\label{subsec:challenges_simulating}
To understand the challenges of simulating middle school students in CPS, we started by exploring a vanilla simulation approach, where we prompted an LLM with sentences specifying the characteristics of the student persona and the context of mathematical CPS, without other architectural system designs. We did not involve real humans during this preliminary exploration, but instead deployed three LLM-simulated students to interact with each other. In \autoref{fig:prelim-exploration} (\autoref{app: prelim-exploration}), we present one example dialogue by running the vanilla simulation and an example produced by \system for a comparison.

From the preliminary exploration, we identify two challenges of simulating middle school students in CPS. 
The first challenge is \textbf{{persona alignment}}, i.e., \textit{aligning an LLM's persona simulation to the authentic persona of real human students}. Middle-school students in real life could struggle with rapid and accurate problem-solving~\cite{lambros2004problem}. They may begin with incomplete or incorrect solutions, gradually refining their reasoning through discussion~\cite{brahier2020teaching}. The process of gradually establishing mutual understanding with peers, interleaved by continual self-reflection, is the key for students to improve their mathematical skills.
However, in our preliminary exploration, a vanilla persona simulation
could not yield such an authentic and fine-grained student simulation. Instead, LLMs often directly gave a close-to-perfect solution, even when prompted to simulate students with low mathematical proficiency. For example, as shown in \autoref{fig:prelim-exploration}, the naively simulated character Alice only verbally claims herself as ``not great at math'', while still giving a correct calculation process during CPS. A similar observation was also found by a concurrent study~\cite{kumar2025can}. Simulating how a student's problem understanding and solving evolve during the collaboration poses a major difficulty in simulating students in CPS. In addition, we observed that vanilla simulation yielded overly lengthy responses, which makes the simulation even more unrealistic. 

The second challenge is \textbf{{conversational procedural alignment}}, i.e., \textit{aligning the overall conversational procedure to an authentic CPS discussion among middle-school students.} While the first challenge concerns individual persona simulation, this second challenge considers interactions among personas and between the human and personas. In practice, a CPS process typically consists of multiple stages, including problem understanding, task division, solution planning, plan execution, etc.~\cite{pisa2018assessment}. However, with a vanilla simulation approach, when we directly exposed the group of virtual students to interact with each other, these virtual students often bypassed the earlier stages and instead initiated the conversation by immediately describing the problem solution. For example, as shown in \autoref{fig:prelim-exploration}, a vanilla simulation starts the conversation directly from problem-solving, whereas the virtual students never have realistic greetings and communications about team collaboration. The three virtual students end the conversation immediately after each of them describes their solution, without confirming shared understanding, argumentations, clarification, and reflection. When deployed to serve a human student, such simulations will limit the participation of the human student, leading to largely ineffective interaction and math learning.

For both challenges, we conjecture that the vanilla simulation fails because LLMs were typically trained in a question-answering style to provide direct and accurate answers. As a consequence, even with carefully designed prompts, they cannot simulate the deliberation of middle school students and the nuances in their interactions. Despite existing exploration of LLM persona simulations, to the best of our knowledge, no prior work has carefully explored both of the two alignment challenges and the solutions to them.

\subsection{Modular Architecture Design}
\label{subsec:modular_arch_design}
To address these two challenges, we propose to design \system with a modular architecture, which includes a \textbf{meta planning} module for organizing the overall conversation and facilitating a smooth multi-stage student discussion, and a \textbf{persona simulation} module for creating and updating individual student personas. The former encourages conversational procedural alignment, while the latter encourages persona alignment.
\autoref{fig: schema} shows an overview of the modular architecture of \system.
Below, we provide both modules in detail, along with our design rationale.
All prompts to the LLM for implementing the module components are shown in~\autoref{Appendix: prompt}.

\subsubsection{Meta Planning}
\label{subsec:meta-planning}
The meta planning module performs meta-level control of the entire conversation to make the conversation remain smooth and closely mirrors the real student interaction process.
It consists of two components.\\

\begin{table*}[h!]
\caption{Collaboration Stages and Associated Dialogue Acts in \system}
\centering
\tiny
\resizebox{.8\textwidth}{!}{
\begin{tabular}{lll}
\toprule
\textbf{Stage} & \textbf{Collaboration Stage} & \textbf{Dialogue Act Candidates} \\
\midrule
Stage 1 & Establish shared understanding &
  \begin{tabular}[c]{@{}l@{}}
  Present a task understanding\\ 
  Ask a clarifying question about task understanding\\ 
  Answer a clarifying question about task understanding\\ 
  Second a task understanding\\ 
  Ask for agreement on a task understanding
  \end{tabular} \\
\midrule
Stage 2 & Establish team organization &
  \begin{tabular}[c]{@{}l@{}}
  Initiate workload division\\ 
  Volunteer to serve a role\\ 
  Second role designation plan\\ 
  Ask for agreement on the role designation
  \end{tabular} \\
\midrule
Stage 3 & Plan actions for problem-solving &
  \begin{tabular}[c]{@{}l@{}}
  Prompt a teammate to join the discussion\\ 
  State an action plan\\ 
  Ask a clarifying question about an action plan\\ 
  Answer a clarifying question about an action plan\\ 
  Second an action plan\\ 
  Ask for agreement on an action plan
  \end{tabular} \\
\midrule
Stage 4 & Execute actions for problem-solving &
  \begin{tabular}[c]{@{}l@{}}
  Execute an action plan and state the execution result\\ 
  Ask a clarifying question about an execution result\\ 
  Answer a clarifying question about an execution result\\ 
  Second an execution result\\ 
  Ask for agreement on an execution result
  \end{tabular} \\
\midrule
Stage 5 & Validate the answer &
  \begin{tabular}[c]{@{}l@{}}
  Reflection on the implication of modeling outcomes\\ 
  Summarize the results
  \end{tabular} \\
\bottomrule
\end{tabular}}
\label{table:stage_and_dialogue_acts}
\end{table*}

\noindent(1) \textbf{Collaboration Stage Monitor}:  
Adapting from PISA 2015's Assessment and Analytical Framework \cite{pisa2018assessment}, we formulate each CPS session as a multi-stage process. The original PISA 2015 CPS framework includes three grand competencies of CPS (i.e., establishing and maintaining shared understanding, taking appropriate action to solve the problem, and establishing and maintaining team organization), each annotated with four fine-grained individual problem-solving processes (i.e., exploring and understanding, representing and formulating, planning and executing, and monitoring and reflecting) sourced from the PISA 2012 individual problem-solving framework \cite{oecd2013pisa}. Since the original PISA framework was designed for \emph{post-hoc} assessment and analysis of students' CPS performance, adaptation is necessary for it to support \emph{proactive} multi-persona simulation at both conversational and individual levels. To this end, we generalize the three grand competencies of CPS in the PISA framework into three stages of CPS, but break down the stage of ``taking appropriate action to solve the problem'' into three more fine-grained stages (Stages 3-5) that incorporate the individual problem-solving processes (i.e., planning actions for problem-solving, executing actions for problem-solving, and validating the answer). Furthermore, ``establishing and maintaining shared understanding'' and ``establishing and maintaining team organization'' are simplified into ``establishing shared understanding'' (Stage 1) and ``establishing team organization'' (Stage 2), whereas the maintenance component is left to be reflected through peer engagement during other stages (e.g., prompting a teammate to discuss in Stage 3, and asking for agreement throughout Stages 1-4). This results in our five-stage CPS formulation shown in \autoref{table:stage_and_dialogue_acts}, which is further engineered with a set of \emph{dialogue acts} \cite{stolcke2000dialogue, li2016user} in each stage, indicating the plausible intents of utterances under the stage, to be used for individual persona simulation.

In our formulation, every CPS dialogue starts with establishing shared understanding, with the collaboration stage monitor dynamically determining whether the discussion should proceed to the next stage (e.g., establishing team organization and planning actions for problem-solving). 
The collaboration stage monitor is implemented by prompting an LLM to decide, based on the dialogue history and the definition of every stage, whether the dialogue should move on to the next stage sequentially. When all stages are completed, we end the session. By introducing the stage formulation, \system is expected to result in much more extended conversations than the vanilla simulation. \\

\noindent(2) \textbf{Next Speaker Control}: 
To facilitate smoother conversation interactions, we also developed a dialogue speaker control module that predicts the next speaker based on the dialogue context. 
For example, if the last response is ``\textit{Hey Bob, you made a mistake...}'', the next speaker is more reasonably to be Bob. The next speaker control module captures this intuition by inputting the dialogue history, and it is prompted to output the virtual student's or human student's name as the next speaker.

When the predicted next speaker is only the user (i.e., the human student), \system awaits the user's input in order to continue the conversation.  
If the user does not perform any valid keyboard action within one minute, a virtual classmate is randomly selected to continue the conversation; the timer resets upon any keystroke.
When the predicted next speaker includes both the user and one or more virtual classmates, the system awaits the user for 10 seconds under the same conditions. 
If the predicted next speaker is solely a virtual student, the persona responds immediately without delay.
These timing thresholds were determined based on insights from our internal testing.
The one-minute timeout accommodates the reading or typing time needed when the user is solely responsible for continuing the discussion.
The 10-second interval balances responsiveness with giving a fair chance for the user to contribute or skip the input when the user or agents can all be the responders.

\subsubsection{Persona Simulation}\label{subsec:character-sim}
\begin{figure*}
    \centering
\includegraphics[width=.85\textwidth]{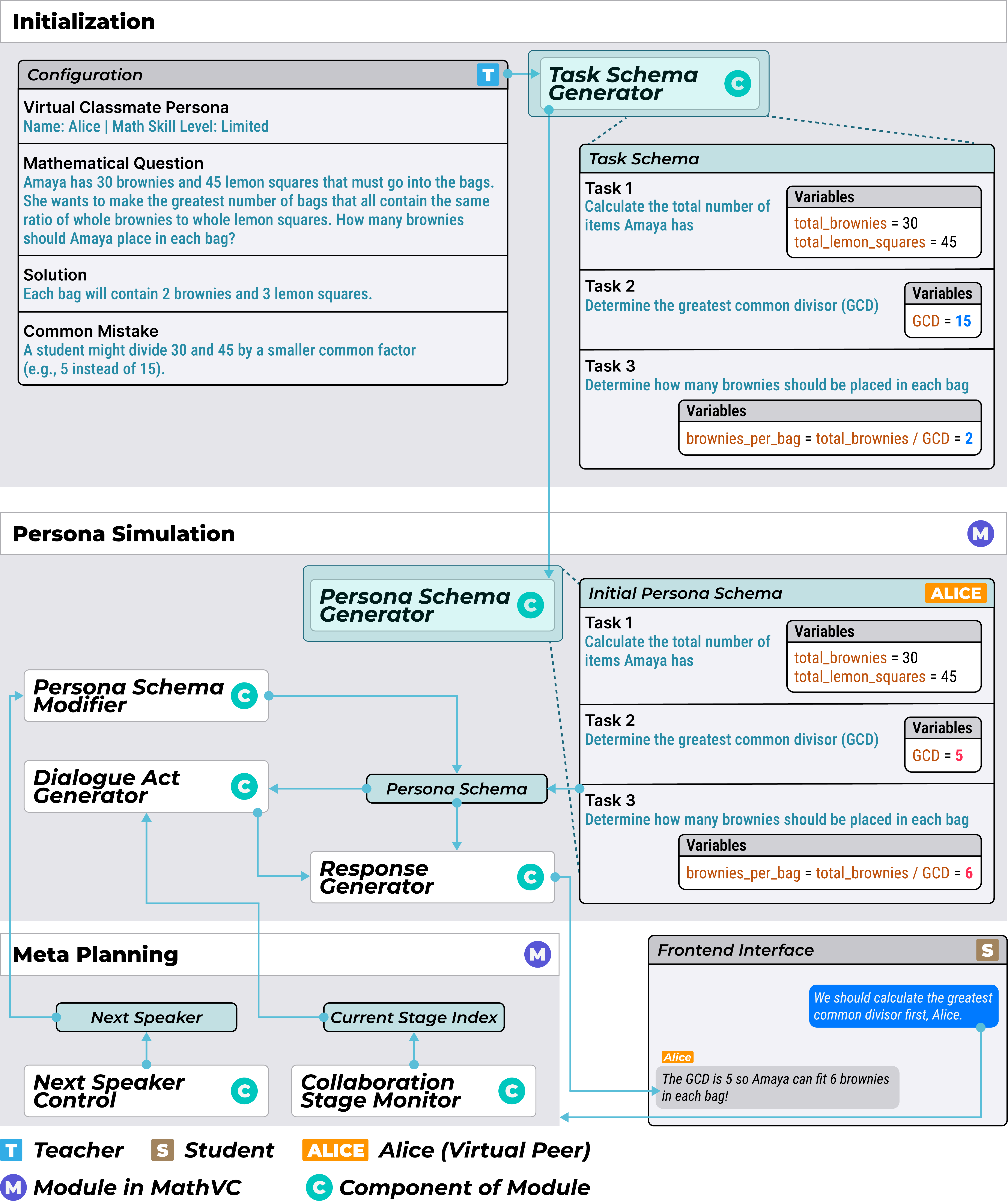}
    \caption{
    Overview of \system's structure. \textbf{(1) Initialization:} The administrator (middle school teacher) first sets up a mathematics problem and a set of virtual peers (e.g., Alice in the example). The \textit{task schema generator} then automatically generates a structured representation of the problem solution to facilitate the CPS.
    \textbf{(2) Persona Simulation:} When a user (middle school student) starts a session on \system, the specified virtual peers are created, where the \textit{persona schema generator} generates a structured representation of each persona's understanding of the problem by potentially altering correct variable values (e.g., changing \textcolor{blue}{15} to \textcolor{red}{5} and \textcolor{blue}{2} to \textcolor{red}{6}) to reflect the persona's mathematical skill level (e.g., Alice's the limited skill). During CPS, the persona schema will also be updated by \textit{persona schema modifier} to reflect the latest knowledge of the persona. Subsequently, the \textit{dialogue act generator} and the \textit{response generator} decide the intent and the content of each virtual peer's next utterance, which is then sent to the frontend for display.
    \textbf{(3) Meta Planning:} The entire CPS dialogue is overseen by the \textit{collaborative stage monitor}, which decides the current stage of CPS, and the \textit{next speaker control}, which selects the next speaker (e.g., Alice).
    }
    \label{fig: schema}
\end{figure*} 
The persona simulation module is to emulate the thought processes and responses of real middle school students. 
For the scope of this project, we focus on simulating the mathematical proficiency level (i.e., advanced or limited) of a virtual peer, in addition to assigning them a name.
Middle-school students in real life may initially produce incomplete or erroneous solutions and refine them iteratively as the discussion progresses.
To simulate the process of gradually establishing mutual understanding with classmates, we propose to use \emph{symbolic schema representations} to describe each virtual classmate's present thought for solving a mathematical problem and seek to dynamically update it to reflect how the virtual classmate changes their understanding throughout the CPS. 

Specifically, we design two schema representations. \textbf{Task Schema} is a symbolic structural representation describing elements that are necessary for solving a given mathematical problem. As shown in \autoref{fig: schema}, a task schema specifies how the mathematical problem can be decomposed into multiple sub-tasks and for each sub-task, what variables should be defined to resolve it. 
In other words, a task schema shows the ground-truth understanding and modeling of a mathematical problem.
We define the schema centering around ``variables'' because they are considered the fundamental elements in mathematical problem solving~\citep{core2010common}. 
The schema is automatically generated by prompting an LLM with the mathematics question and the ground-truth answer. 

Built on top of the task schema, we further devise \textbf{Persona Schema}, a symbolic structural representation showing a student persona's understanding and modeling plan on the given mathematical problem. A persona schema is a task schema injected with errors that could be made by real human students. 
We consider a wide range of errors, including misunderstanding a sub-task or even omitting it, as well as any variable mistakes resulting from miscounting, incorrect modeling, or calculation errors. In practice, the errors could be specified by the teacher setting up \system based on their teaching experience and the learning objectives. 
In our human study (\autoref{subsection: study}), we first enquired the LLM about common mistakes for each mathematical problem used in the study, and then manually verified them with the mathematics education expert in our team, including making necessary modifications to make the mistakes more representative. 
Throughout the conversation, the student persona will gradually update their persona schema to reflect their dynamic thought process, which serves as a foundation for achieving the persona alignment.

To simulate the student persona, the persona simulation module in \system includes four components.\\

\noindent(1) \textbf{Persona Schema Generator:} 
We first provide a specific question and its corresponding answer and prompt LLM to generate the associated subtasks and the variables involved in each subtask, forming the \emph{task schema}. 
For simulated students who are set to perform well, the task schema is directly used as the persona schema. 
For simulated students with weaker math skills, we collect common errors for each problem and generate a persona schema with an error selected from the provided common errors by the LLM. 
As a result, each simulated student character possesses a unique character schema that reflects their initial thinking before engaging in discussion with the real student.\\

\noindent(2) \textbf{Persona Schema Modifier:}
During real-life student discussions, a human student's thought for a given question often evolves as the conversation progresses. 
We simulate these dynamics by continuously modifying the persona schema based on the dialogue history. 
When the student persona is selected as the next speaker, we prompt the LLM to determine whether the current persona schema requires modification.
The decision is typically based on whether the ongoing conversation leads to a change in the student persona's mathematical understanding, for instance, if the student corrects a mistake made by the student persona. In such cases, the persona schema needs to be updated to reflect the new understanding. To achieve this, the persona schema modifier generates a JSON object in the format \texttt{\{variable\_name: updated\_value...\}}, specifying the variables to be changed and their new values. This updated information is then used to revise the original persona schema accordingly. 
\\

\noindent(3) \textbf{Dialogue Act Generator:}  
After obtaining the latest persona schema, the dialogue act generator is leveraged to pick a ``dialogue act'' indicating the intent of their response. 
Dialogue acts are communicative functions of utterances  (e.g., asking a question, explaining a concept, etc.)~\cite{searle1969speech,navarretta-paggio-2020-dialogue}, and are widely used in traditional dialogue systems~\cite{stolcke2000dialogue, li2016user, zhao2017learning}.
We found that the use of dialogue acts could increase the response diversity.
For instance, instead of repeatedly responding with generic confirmations like ``I agree,'' the system can choose more specific acts, thereby enriching the interaction and encouraging deeper engagement. In addition, by specifying a concrete response intent, a dialogue act encourages more focused and concise responses, which greatly mitigates the issue of lengthy responses observed with the vanilla simulation approach.
In our application, we define a set of dialogue acts for each CPS stage, such as ``present a task understanding'' and ``ask for a clarifying question about a task understanding'' when the conversation are in the stage of establishing a shared understanding, as shown in \autoref{table:stage_and_dialogue_acts}.
Although we provide a set of candidate dialogue acts, we still require the dialogue act generator to generate a context-specific act rather than simply selecting from the list, in order to offer more specific guidance.
The dialogue act generator is tasked to \emph{generate} a dialogue act (e.g., \textit{``ask Bob a clarifying question about...''}) based on the persona schema, the dialogue history, and the set of dialogue acts for the current stage. 
\\

\noindent(4) \textbf{Response Generator:} 
Once the dialogue act is decided, the persona's response is generated based on the generated dialogue act and the conversational context.
To encourage language styles closer to middle school students', we also include instructions to prompt the LLM to generate short and colloquial responses.

\subsection{Implementation of \system}
\label{subsec:implementation}
We implemented \system with distinct frontend and backend components. The frontend, a user-facing web application, was built using TypeScript~\cite{typescript} and React~\cite{react}, and the backend was implemented with the Gentopia agent framework~\cite{xu2023gentopia}.
Key frontend interface elements and backend components are illustrated in \autoref{fig: VC} and \autoref{fig: schema}. In the current instantiation of \system, we use OpenAI GPT-4o--2024-05-13, the state-of-the-art LLM by the time of the study, to power the system's backend. The LLM is modular and can be readily replaced by other models.

\noindent\textbf{System Deployment.}
Both the backend and the frontend applications were deployed on a single server. Communication between the user's browser and the backend logic was handled in real-time using WebSocket connections. 



\noindent\textbf{Session Configuration.} \system allows administrators (e.g., teachers) to configure accounts for users (e.g., students) and assign tailored mathematical problem sets to distinct users or groups. Each mathematical problem can then be configured with its correct answer(s) and associated common misconceptions, which are subsequently used by \system to generate the corresponding task schema and the persona schema for each virtual peer, 
as generally illustrated in \autoref{fig: schema}. Furthermore, \system enables the assignment of different virtual peers to particular mathematical problems for each user.


\noindent\textbf{User Interface Components.} Users initiate their session by logging into \system with credentials configured by their administrators. The primary user interface consists of a header that displays the current mathematical problem and allows navigation between different assigned problems. The main panel is a chatbot that enables users to interact with their virtual peers for mathematical CPS (see \autoref{fig: VC} for visual details).

\noindent\textbf{Data Management.} All interaction data, including chat logs and problem-solving steps, was securely stored in a backend database. To facilitate the post-session interview, a temporary copy of the current session's interaction data was also cached in the user's browser storage. The temporary data was automatically deleted when the user closed the browser tab or window, to ensure data persistence only on the secure backend.

\section{Evaluation Method}
Upon approval from our Institutional Review Board (IRB), we conducted an evaluation study with 14 middle school participants from the United States. Below, we describe our participant recruitment and overview, the study procedures, and data analysis methods. 


\subsection{Participant Recruitment and Overview}
We used multiple strategies to reach families with children in grades 6–8 (i.e., middle school students). Invitations were distributed by contacting colleagues and friends who are parents in this age group and posting recruitment flyers on social media platforms such as Nextdoor. Interested parents were asked to complete an initial participant form. The participant form asked about their child's demographic information, including grade level, age, gender, race and ethnicity, and available time slots for scheduling the study session. 


In total, 14 participants from grades 6–8 participated in our study (also see \autoref{tab:demographic}), including five from grade 6, four from grade 7, four from grade 8, and one advanced student below grade 6. Among participants, six identified as male and eight as female. Participants' self-reported racial and ethnic identities included Asian or Asian American (n=3), Black or African American (n=2), White or European (n=6), and three preferred not to specify.


\begin{table*}[ht]
\centering
\caption{Demographic information of participants.}
\label{tab:demographic}
\small
\begin{tabular}{llllll}
\toprule
\textbf{PID} & \textbf{Grade} & \textbf{Age} & \textbf{Gender} & \textbf{Race \& Ethnicity} \\
\midrule
P1 & Grade 7 & 12 & Male & Asian or Asian American \\
P2 & < Grade 6 (from a talented class) & 10 & Male & White or European \\
P3 & Grade 7 & 13 & Female & Asian or Asian American \\
P4 & Grade 7 & 12 & Female & White or European \\
P5 & Grade 8 & > 13 & Female & Asian or Asian American \\
P6 & Grade 8 & 13 & Female & Prefer not to say \\
P7 & Grade 6 & 11 & Female & Prefer not to say \\
P8 & Grade 8 & 13 & Male & Black or African American \\
P9 & Grade 6 & 11 & Male & Prefer not to say \\
P10 & Grade 8 & 13 & Male & Black or African American \\
P11 & Grade 6 & 11 & Female & White or European \\
P12 & Grade 7 & 11 & Female & White or European \\
P13 & Grade 8 & 13 & Female & White or European \\
P14 & Grade 7 & 13 & Female & White or European \\

\bottomrule
\end{tabular}
\end{table*}

\subsection{Study Procedures}
\label{subsection: study}
We first ran a series of pilot sessions within our research group to finalize the mathematical problem sets for participants and address usability issues in \system.

\noindent\textbf{Customized Question Selection \& Rationale:} Our question selection process began with sourcing grade-appropriate mathematical problems from the Illustrative Mathematics\footnote{\url{https://illustrativemathematics.org/}} and Math-Mapper\footnote{\url{https://www.sudds.co/}} websites, suggested by our mathematical education domain expert in the team. We then manually curated three questions for each grade, ensuring a spectrum of difficulty. Selected questions underwent slight textual modifications to enhance clarity. The resulting grade-specific problem sets were then finalized based on insights gained from the pilot sessions. Once established, we looked into each participant's screening survey to tailor their evaluation study materials by creating a \system profile with three pre-determined mathematical problems, corresponding to their grade. A detailed list of questions used in our study can be found in the appendix (\autoref{tab:ratio_questions_full}).

\noindent\textbf{\system Virtual Peer Setup:} 
We included two virtual peers, Alice and Bob, in our evaluation study. 
Alice was simulated with limited mathematical skills using an error-injected persona schema, while Bob was simulated with advanced skills using an error-free persona schema. 
We intentionally limited the number of virtual peers to two to maintain simplicity, clarity, and manageability of interactions for participants, which allows for focused observations of students' collaborative interactions with distinctly characterized virtual personas.

\noindent\textbf{Overall Study Procedures:} Each study took approximately 60 minutes, including four parts: 1) onboarding ($\approx$ 5 min), where the research team introduced participants to \system and its functionalities, 2) problem-solving using \system ($\approx$ 30 min), where students tackled mathematical problems with their virtual peers, completing two or three problems based on their pace within the timeframe and 3) follow-up semi-structured interview ($\approx$ 25 min) that allowed us to further understand participants' interaction experiences, perceptions of authenticity, engagement, and suggestions, and 4) an exit survey that included seven Likert scale questions that examined students' sense of reaching a shared solution, their own contribution and planning, the realism of virtual peers, and how closely the session in \system resembled typical in-person group work for middle school settings. At the end of the study, each participant was compensated with a \$30 gift card.

\subsection{Data Collection and Analysis}

Our collected data comprised three primary components: logs of student interactions with \system, interview conversations, and responses to the exit survey.

We began our analysis by calculating descriptive statistics for all quantitative variables, reporting means and medians for continuous data. For the interview conversations, we conducted thematic analysis guided by \textit{General Inductive Approach}~\cite{thomas2006general}. Initially, one lead author conducted a thorough review of the transcripts and generated low-level codes by labeling concepts. These low-level codes were subsequently clustered into broader thematic categories, such as \textit{Collaborative Dynamics} and \textit{Engagement \& Motivation}. Throughout the analysis, the research team held regular meetings to review and refine the emerging themes. A codebook outlining major themes is provided in \autoref{Appendix:codebook}. 

\section{Findings}
Our findings combine analysis from students' conversational logs in \system, their exit survey results, and semi-structured interviews to reveal insights into participants' experiences with \system. Below, we first present descriptive statistics of student conversation using \system and students' exit survey results showing their positive attitudes. We then describe results related to participants' perceived authenticity of \system in \autoref{subsec:authenticity}, as well as their views on the role of \system in supporting collaborative mathematical problem-solving (\autoref{subsec:5-2} to \autoref{subsec:confidence}).

\label{subsec:rq1}
 
We collected 40 dialogue sessions from 14 participants, each focused on solving a single mathematical problem collaboratively with virtual peers in \system. Our data comprises 695 messages in total, averaging 15.5 conversational turns per problem, with 295 messages sent from students. Messages from \system were further divided between its two personas: 166 generated by \textit{Alice} and 133 by \textit{Bob}.  

\textbf{Exit Survey Results Showing Overall Positive Attitude Towards \system.}  
As shown in \autoref{tab:survey_stats}, participants reported high levels of constructive interaction with the simulated classmates (Mean = 4.36, Median = 5) and strongly agreed that the group successfully reached a shared understanding or solution (Mean = 4.43, Median = 5). Furthermore, they felt they could understand the math problem and work effectively within the virtual environment (Mean = 4.21, Median = 5), reinforcing the qualitative feedback about the system's helpfulness and engaging nature in facilitating collaborative mathematical problem-solving.

\begin{table*}[ht]
\centering
\footnotesize
\caption{Exit survey responses from participants. All questions used a 1–5 Likert scale, where 1 indicates strong disagreement and 5 indicates strong agreement.}
\label{tab:survey_stats}
\begin{tabular}{p{0.7\textwidth}l}
\toprule
\textbf{Exit Survey Question} & \textbf{Response Statistics} \\
\midrule
I could understand the math problem and work effectively within the virtual classroom environment. & 
\text{Mean=4.21, Median=5, Min=3, Max=5} \\
\midrule

I interacted constructively with the simulated classmates (e.g., asking questions, building on their ideas). & 
\text{Mean=4.36, Median=5, Min=2, Max=5} \\
\midrule

The group and I were able to come to a shared understanding or solution to the problem. & 
\text{Mean=4.43, Median=5, Min=3, Max=5} \\
\midrule

I took responsibility for my role and contributed to completing the math task efficiently. & 
\text{Mean=4.29, Median=4, Min=3, Max=5} \\
\midrule

I planned my steps and understood how each action (mine and my classmates') contributed to solving the problem. & 
\text{Mean=4.36, Median=4, Min=4, Max=5} \\
\midrule

I believe chatting with a computer-aided peer will feel similar to talking with a real student. & 
\text{Mean=3.86, Median=4, Min=2, Max=5} \\
\midrule

I expect that working with virtual classmates is similar to the group work I do when solving problems. & 
\text{Mean=4.21, Median=4, Min=2, Max=5} \\
\bottomrule
\end{tabular}
\end{table*}

\subsection{Mixed Perceptions of The Sense of Authenticity of \system}
\label{subsec:authenticity}
Our findings show that middle school students in our study judged the ``realness'' (or the sense of authenticity) of their virtual peers on three aspects, including \textbf{collaborative behaviour} (e.g., did the agents start, share, and revise ideas like real peers?), \textbf{human-like fallibility} (e.g., did they occasionally make errors and then correct them?), and \textbf{social ``voice''} (e.g., did they sound and act like middle-schoolers?). 


\subsubsection{Where the Agents Felt Real among Middle Schoolers} 
Some participants (n=6) felt their interactions with virtual peers within \system closely mirrored their real-world classroom experiences. Particularly, participants appreciated the \system's \textbf{familiar openings} (e.g.,  how the virtual peers launched discussion in familiar ways) and human-like fallibility (e.g., how they made and fixed mistakes like real students); both of which are known triggers of productive peer talk in middle-school mathematics.

Participants frequently used terms such as ``pretty much the same'' or ``pretty similar'' to describe the resemblance. P3, for example, recognized a typical opening pattern shown in \system when comparing with a real-world classroom setting:
\begin{quote}
    \emph{``It was pretty much the same, because the way they start the conversation is a lot like people in a regular class when they share their answers and collaborate.'' (P3)}
\end{quote}

For P3, the virtual peers' method of initiating discussion felt natural and recognizable, contributing positively to the perceived authenticity. In classroom discourse, such initiation moves are critical as they set a norm of shared ownership and predict deeper explanation later, as prior literature described \cite{cazden2015study, zhu2022discourse}. The fact that P3 instantly recognized the pattern suggests that \system reproduced those openings convincingly.

Additionally, the sense of realism of the virtual classroom was also amplified by the virtual peers' designed capacity to make errors, as noted by P6:

\begin{quote}
    \emph{``Sometimes it would like make mistakes... it just makes the bot feel a little more human because people make mistakes.'' (P6)}
\end{quote}

P6's explanation links virtual peers' imperfection to perceived human-like qualities, suggesting that flawless AI might feel less authentic in a peer simulation context. In other words, the fallibility makes the interaction more relatable and believable, which mirrors the reality that peers are not always correct. 

Authenticity was strengthened even further when the virtual peers critiqued the student’s work, an interaction that middle-schoolers recognise as ordinary peer feedback. P7 illustrated: 

\begin{quote}
    
    \emph{``They (virtual peers) actually tell you that your response is wrong, not that it's right, and they tell you what you have done wrong also, which is pretty similar [to real experience].'' (P7)}

\end{quote}
By describing the peers with the human pronoun \emph{they} and attributing evaluative agency (``tell you what you have done wrong''), P7 framed the \system as full conversational partners rather than automated tools. 

Such personification was also reflected in the students' direct communication with their virtual peers during the tasks. For example, P2 responded directly to a virtual peer's suggestion by typing, \emph{``I like your thinking.''} The explicit affirmation of the virtual peer's cognitive contribution (``your thinking'') demonstrates that students often interacted with virtual peers as if they were engaging with actual human classmates, capable of reasoning and deserving of social acknowledgment.

Through interactions with \system, the usages of pronouns and direct conversational acknowledgments further reveal the perceived authenticity of the simulation. It indicates that the virtual peers in \system successfully mimicked key aspects of human peer interaction, leading students to engage with them in socially meaningful ways, treating them as human partners.

\subsubsection{Where the Illusion Broke Down} 
Although many students in our study credited \system with ``human-like'' behavior, a couple of participants also pointed to cues that made the agents feel \textit{off}. The \textbf{adult-sounding language} highlights the potential design enhancements. 

We found that middle-school talk tends to be brief, casual, and peppered with slang, whereas the virtual peers sometimes replied in full, polite sentences. P1 contrasted the two styles:
\begin{quote}
    \emph{``My day-to-day classmates ... would just keep it really normal and casual, and they would honestly keep it really short. They wouldn't try to explain it like really, really long.'' (P1)}
\end{quote}


P1 indicated that the virtual peers' formality and well-structured detail clashed with students' brief, slangy talk in real life, so \system's dialogue felt less authentic to some. The perceived lack of casualness made \system's dialogue feel less authentic for some. P9 echoed the sentiment, noting the virtual peers' words \emph{``are formal and not kids''} and lacked the typical peers' use of  \emph{``shorter sentences and slang''}.

\takeaway{Overall, students in our study perceived the high authenticity of \system's virtual peers. Participants appreciated \system's ability to mimic key aspects of peer collaboration, including making mistakes and offering corrections, which enhanced realism for some. Yet, some participants also pointed out a few aspects to be improved to enhance the ``realness'' or peers in terms of the communication style and the need for less formality. }

\subsection{Socio-Emotional and Motivational Engagement via Peer Interaction}
\label{subsec:5-2}

Our participants emphasized the critical role of socio-emotional and motivational dimensions when engaging in collaborative mathematical problem-solving with \system. Students valued aspects of their interaction experiences with \system, such as \textbf{reduced cognitive burden, diverse perspectives supported by different LLM personas, friendly competition, and immediate feedback}, which collectively enhanced their motivation and persistence in math problem solving.

\subsubsection{Reducing Cognitive Load through Collaborative Atmosphere.} 
Several students in our study acknowledged how collaboration reduced cognitive demands and boosted intrinsic motivation. P2 highlighted, \emph{``It feels like the problem is easier when I'm collaborating.''} Such perceptions align with collaborative learning literature, which indicates that peer interactions help distribute cognitive load and support sustained effort~\cite{kirschner2018cognitive}.  

Additionally, P4 highlighted the value of shorter, peer-like conversational turns in contrast to teacher-led instruction, 
\begin{quote}
\emph{``If it was like (\system) versus like a teacher, it would be less engaging. Sometimes I feel like if it's like a teacher up there, it just sounds like someone's talking for a long time... I feel like that sometimes is one of the reasons that people just kind of stop listening. '' (P4)} 
\end{quote}
P4's experience illustrates how interactive, brief exchanges characteristic of \system mitigate cognitive overload, which could foster continuous attention and participation.

\subsubsection{Scaffolding Deeper Reasoning through Diverse Perspectives}
\label{subsubsec:scaffolding_diverse_perspectives}
Participants noted that differing viewpoints provided by the virtual peers acted as cognitive scaffolds, encouraging deeper engagement and critical evaluation. Specifically, many students in our study observed that virtual peers sometimes disagreed or used different strategies, indicating a sense of \textit{heterogeneity}. Such heterogeneity, reflective of authentic classroom dynamics, offered crucial scaffolding by prompting students to elaborate on and critically assess multiple problem-solving pathways. Indeed, prior research suggests that diverse perspectives in collaborative settings enhance conceptual understanding by facilitating comparative reasoning and strategy adaptation \cite{cavazos2024greater, chandra2015collaborative}. For example, 
P8 explicitly benefited from these differing viewpoints:

\begin{quote}
    \emph{``I did notice that they had two different opinions when it came to the math problem... It did help me think through different ways, that I could get the problem right. '' (P8)}
\end{quote}

Similarly, observing alternative problem-solving methods led P5 to consider adopting potentially more efficient strategies, which seems to enhance their own mathematical understanding:

\begin{quote}
    \emph{``I saw, like, maybe Alice use a different math way, I may try and do the same math, like the way that she did. So ... maybe I would use the same method that Alice did because ... it's faster and easier.'' (P5)}
\end{quote}

Moreover, the variation in virtual peer skill levels provided a context for reciprocal teaching, a powerful scaffolding mechanism established in collaborative learning research for reinforcing conceptual clarity \cite{rosenshine1994reciprocal}. P3 explicitly highlighted this benefit:

\begin{quote}
    \emph{``Because it (\system) made it more diverse. It made it better because you could help the classmates, too, while learning for yourself. '' (P3)}
\end{quote}

P3's quote suggests how diverse perspectives in \system not only could scaffold individual understanding but also foster reciprocal scaffolding opportunities, which enables students to consolidate their knowledge by teaching and assisting peers.


\subsubsection{Friendly Competition Driving Proactive Participation}
The interaction in \system sparked a sense of friendly competition for some students, adding another layer to their motivation. P11 expressed this explicitly:

\begin{quote}
    \emph{``I'll always find a way to beat them. Always make sure I want to beat them. I don't want them to give the answers before me, so I feel it's motivated me.'' (P11)}
\end{quote}

P11's desire to \textit{beat} the virtual peers introduces a competitive element present within collaborative problem-solving and peer learning environments, where `beating' does not necessarily imply hindering, but rather using teammates as a ``benchmark'' to push personal performance and demonstrate competence within the shared task. 
Consequently, it stimulated students to proactively propose potential problem-solving methods and attempt to lead the discussion. For example, P5 took the initiative early on to set the direction, stating, \emph{``it looks like a multiplication problem so maybe we should focus on that''} before the virtual peers offered a detailed path. Similarly, P6 demonstrated leadership by managing the collaborative flow, prompting a virtual peer by asking, \emph{``Alice, what do you think would be the first step in solving this?''} Students' drive to contribute first or guide their collaborative process showcases how the competitive dynamic in \system could translate into their motivation in problem-solving tasks.


\subsubsection{Immediate, Targeted Feedback Promoting Persistence.} 
Participants frequently emphasized the motivational impact of immediate and targeted feedback from virtual peers, highlighting the value of \system over more passive or delayed responses common in other educational technologies. For example, P7 contrasted their experience explicitly with ChatGPT:

\begin{quote}
    \emph{``On ChatGPT, for example, you say, I'm stuck on this problem. I need help. Can you help me? Usually, [it] just tells you `Oh, this would be the answer.' And then you don't really learn anything. But collaborating with Alice and Bob, you can see that you can build on what you know, and you can build on what you don't know... And then you can solve the problem by yourself except from just an answer from online. '' (P7)}
\end{quote}

P7’s reflection emphasizes how collaborative knowledge construction within \system, rather than mere provision of answers, fosters deeper cognitive and emotional investment. Similarly, P14 highlighted this targeted, immediate feedback as exceeding typical classroom peer support:



\begin{quote}
    \emph{``This (mathematical CPS in \system) was honestly more helpful than it would be in school, because I feel like a lot of people didn't really offer suggestions on how we could solve it better (in school).'' (P14)}
\end{quote}

P14's statement underscores a crucial distinction: interactions within \system often provided instant, constructive guidance focused on improving the problem-solving strategy, provided by the ability of LLMs. While in typical school group work, such targeted, improvement-oriented feedback from peers might be infrequent or absent without teacher intervention. The interactions in \system, therefore, not only offered immediacy but also a level of actionable, pedagogical input that students found lacking in their conventional peer collaborations.


\takeaway{Students in our study described \system's collaborative environment as enhancing their socio-emotional engagement and motivation. Interaction with diverse virtual peers reduced cognitive burdens and facilitated reciprocal teaching, while friendly competition encouraged active participation. Furthermore, immediate and targeted feedback fostered a deeper cognitive and emotional investment. Collectively, these findings show \system's potential as an engaging learning tool for promoting CPS engagement in mathematics.}

\subsection{Building Confidence through Scaffolding and Articulation}
\label{subsec:confidence}
The enhanced engagement and motivation fostered by collaboration in \system often translated into increased confidence in students' mathematical abilities. Participants appreciated \system's \textbf{supportive scaffolding} through assistance from virtual peers to help themselves enhance confidence; meanwhile, they gained their metacognitive confidence through \textbf{articulation and error handling} enabled by \system. 

\subsubsection{Supportive Scaffolding through Virtual Peers Assistance}
The enhanced engagement and cognitive scaffolding observed through diverse peer interactions in \system (see ~\autoref{subsubsec:scaffolding_diverse_perspectives}) often translated into increased confidence in students' mathematical abilities. Participants frequently reported feeling more assured after successfully navigating problems with \system's help, receiving validation, or articulating their own understanding. Such a boost often stemmed from perceiving the virtual peers as effective scaffolds, as P2 stated:

\begin{quote}
    \emph{``With Alice and Bob. I feel more confident ... feel like they helped me quite a lot ... keep me on track.'' (P2)}
\end{quote}

P2's reflection indicates that the virtual peers provided necessary support (\emph{``helped me quite a lot''}) and guidance (\emph{``keep me on track''}), enabling the students to overcome difficulties and feel more capable in their mathematical thinking. The perception of \system as a reliable support system capable of bridging knowledge gaps was echoed by P6, who anticipated feeling \emph{``more confident, especially if some of the problems I don't necessarily know how to solve''} could be approached with \system's assistance.

\subsubsection{Metacognitive Confidence through Articulation and Error Handling}
Beyond supportive scaffolding, \system also promoted metacognitive confidence (i.e., students' belief in their capacity to reflect on, monitor, and control their cognitive processes)~\cite{luttrell2013metacognitive} through error handling and opportunities for articulation. Metacognitive confidence is crucial in mathematical problem-solving as it supports learners' ability to self-assess understanding, recognize misconceptions, and effectively adjust strategies to overcome challenges, as prior literature highlights~\cite{izzati2018influence}. In our study, participants frequently reframed errors positively, interpreting virtual peers' corrections not as failures but as learning opportunities. 
P7 explained how they reframed the experience of making an error:

\begin{quote}
    \emph{``I think it gives you a bit more confidence since when you get something wrong. Usually, you're like, `Oh, wait! I did this wrong. This is really easy. I could have done it.' And then you like, want to do more work so you can move on to like a better level.'' (P7)}
\end{quote}

P7's perspective highlights how the system encourages a growth mindset; errors become diagnostic tools, prompting further effort rather than sources of discouragement. 


Students also noted that \system prompted them to explain their thinking, whether instructing the AI or correcting its errors, thereby organizing their reasoning and strengthening understanding and confidence. As P5 noted:
\begin{quote}
    \emph{``I think I became a bit more confident, like explaining [to] other people how to solve math problems and stuff.'' (P5)}
\end{quote}

P5's experience suggests the metacognitive benefit of articulation as teaching or explaining reinforces one's own grasp of the material, thereby building confidence, which is sometimes amplified when students identify and correct virtual peers' mistakes, further validating their own knowledge. For some, like P13, the overall experience within the virtual environment felt more conducive to confidence than interactions with real classmates: \emph{``I feel more confident working in virtual class classmate than my real classmates''}, suggesting that the controlled, supportive, and responsive nature of the AI interaction created a psychologically safer space for students to engage with mathematics, thereby positively impacting their self-efficacy within the context of collaborative learning.

\takeaway{In short, our participants found that \system enhanced students' confidence through supportive scaffolding and metacognitive experiences. The virtual environment embedded in \system encouraged a growth mindset, reduced anxiety about making errors, and enhanced students' belief in their own problem-solving capabilities.}

\section{Discussion}
Our study results reveal how students interacted with LLM-based tools in collaborative mathematical problem solving and provide insights into the collaboration dynamics. Based on our findings, we discuss simulating peers via LLM-based technologies for collaboration, and advocate for more work that leverages theoretical frameworks for collaborative problem-solving in mathematics with LLM personas. We also provide design implications, discuss generalizability that goes beyond middle school contexts, and ethical considerations for collaborative LLM agents. 

\subsection{Simulating Peers via LLM-based Technologies}
\label{sec: peer authenticity}
Our findings reveal that participants valued the authenticity of \system's virtual peers primarily along dimensions of collaborative behavior, human-like fallibility, and social voice. Students recognized realistic collaborative behaviors,  particularly in how the virtual peers initiated discussions and engaged in productive error-making and correction--which positively suggests \system's sense of realism. These findings shed light on the importance of simulating key interactional norms and conversational structures found in actual classrooms. 

However, some participants pointed out that the virtual peers in \system sometimes sounded overly formal, polite, or repetitive. Instead, they would prefer a more casual, slang-rich, and concise style typical among middle-schoolers. The observed discrepancy observed here might be related to the persona response generator component, where we instructed the LLM to mimic middle-school students' communication styles, but might not sufficiently constrain or adapt the model's default tendencies toward structured and verbose outputs. Moreover, the persona simulation, although dynamically updating personas' cognitive states and knowledge levels, might not capture all subtle linguistic cues characteristic of middle-school peer interactions. Such linguistic nuances often depend on implicit social understanding and group-specific cultural practices, which can be challenging for general-purpose LLMs to replicate.

Our findings also highlight important considerations for future designs of LLM personas for collaboration, suggesting that enhancing authenticity requires careful attention not only to content accuracy and error simulation but also to social authenticity in linguistic style and communication patterns.
Such limitations in faithfully simulating specific groups or individuals in linguistic style are recognized as one of the greatest challenges in LLM persona simulation.
For example, \citet{samuel2024personagym} identified language habits as the most difficult challenges in simulating personas.
Similarly, \citet{taubenfeld2024systematic} found that when LLM agents are prompted to engage in debates from specific political perspectives, they often fail to simulate diverse communicative behaviors flexibly but usually default to responses shaped by the model’s inherent beliefs.
In the context of student simulation, prior research indicates that even the most advanced LLMs struggle to follow counterfactual instructions when simulating lower-performing student personas~\cite{kumar2025can}, and LLMs struggle more in responding to children compared with responding to adults~\cite{french-etal-2024-aligning}.
Several possible reasons for the misalignment in simulation have been identified, including distributional biases in pretraining data~\cite{shao2023characterllm} and the fact that LLMs are generally optimized to behave and communicate in the style of helpful assistants, rather than adopting the diverse patterns of real human users~\cite{chen2024persona}.

One potential direction is to employ an open-weight LLM and fine-tune or post-train it on youth dialogues~\cite{shao2023characterllm,zhou2023characterglm}.
However, a critical challenge here is the lack of middle school classroom datasets that include mathematical CPS dialogues.
Existing educational resources cover adjacent areas, such as teacher–student tutoring sessions~\cite{macina2023mathdial,liang2024mathchat}, classroom transcripts~\cite{suresh2022talkmoves}, and child‑caregiver interactions~\cite{liu2024benchmarking}, but they do not target peer CPS, which calls for community effort to collect more for educational applications that support CPS.
\subsection{Leveraging Theoretical Frameworks to Analyze and Enhance LLM Personas in Mathematical CPS}
A critical aspect of successful CPS lies in establishing shared understanding and effectively coordinating reasoning steps among collaborators~\cite{hansen2022students}. In mathematics education, it involves clearly articulating problem-solving strategies, critiquing peers' approaches, and adapting to diverse solutions~\cite{evans2014developing}. Our findings indicated that interactions within \system supported key CPS behaviors among students, including articulating reasoning, critiquing peers' ideas, and considering diverse solution strategies. These behaviors align with core mathematical competencies outlined in the Standards for Mathematical Practice (MP)~\cite{CCSSM2010}, a widely recognized framework describing essential practices mathematics educators should foster. 

We reflected on how our findings and interactions with middle school students in our study might be aligned with MPs. For example, when virtual peers presented differing viewpoints or made errors (\autoref{subsec:authenticity}), students were prompted to make sense of problems and persevere in solving them (MP1) and construct viable arguments and critique the reasoning of others (MP3) as they evaluated virtual peers' suggestions or explained their own thinking. The need to articulate steps clearly to virtual peers, either to guide or correct them, supports attending to precision (MP6) in communication. Furthermore, exposure to potentially different strategies from virtual peers encourages students to look for and make use of structure (MP7) and consider multiple solution paths. \system's role as a collaborator, rather than just an answer key, transforms the interaction into a space for practicing these essential mathematical habits of mind. 

Although our initial system design did not explicitly incorporate the MP framework, students' interactions exhibited significant alignment with these standards. We consider that such alignment indicates a promising, yet currently underexplored opportunity: integrating established educational theories and frameworks, such as MP standards, into the design and evaluation of LLM-powered educational personas.

To more rigorously leverage such alignments, we propose two complementary future research directions. First, leveraging theory-driven frameworks such as MP standards to systematically analyze interaction data can help researchers and educators gain deeper insights into how virtual peers with different personas facilitate mathematical practice and proficiency. Second, incorporating MP or similar theoretical frameworks into future designs of LLM-based personas would ground system functionalities in well-established educational theory, which could guide interaction behaviors toward fostering key collaborative and mathematical skills. Currently, much research in LLM personas lacks grounding in learning theories or educational standards, as prior work has argued~\cite{tseng2024two}. Thus, integrating theory-based frameworks is crucial to systematically enhancing the effectiveness, authenticity, and pedagogical validity of future agent designs.

\subsection{Design Implications, Generalizability, and Ethical Considerations for Collaborative LLM Agents}
Our findings carry a few implications for the design of future LLM-based collaborative learning environments, alongside ethical considerations to be addressed for future work.

First, the mixed perceptions of authenticity expressed by our participants highlight a potential design tension. While functional realism (e.g., making errors) enhanced engagement, the overly formal tone in the dialogue detractd from perceived authencity, to some participants. 
Future design could strategically select which aspects of peer interactions to authentically replicate (e.g., natural conversational dynamics and error-making), and which to optimize for pedagogical clarity (e.g., explanatory feedback). Designers might also implement user-customizable or adaptive conversational styles to better align virtual peer interactions with specific student populations.

Second, students in our study valued immediate, actionable feedback and interactions with virtual peers who exhibited diverse problem-solving strategies and varying proficiency levels. These findings emphasize the importance of designing collaborative LLM agents capable of providing detailed, explanatory feedback that moves beyond simple correctness judgments. The heterogeneity within multi-agent systems like \system has the potential to enrich the learning experience by closely mirroring real-world classroom group dynamics. From a CHI perspective, designing diverse virtual peer interactions can foster richer, more collaborative environments that better represent the range of abilities and perspectives encountered in actual educational settings. Such dynamics facilitate deeper cognitive engagement and critical reflection, as students actively negotiate differences, reconcile diverse strategies, adapt their reasoning, and benefit from reciprocal peer teaching~\cite{zhou2024disagreeing}. Future research may consider systematically exploring the creation and evaluation of heterogeneity in virtual collaborative platforms, potentially drawing on established theories of collaborative learning and group dynamics.



Regarding \textbf{generalizability}, although our study specifically targeted middle-school mathematics contexts for the scope consideration, the underlying design principles and LLM-driven collaborative interactions could hold promise for broader scenarios. For elementary contexts, we suspect that adjustments may be necessary to simplify agent dialogues and scaffolding mechanisms, matching younger learners' developmental and cognitive needs. For high school contexts, future design might require more complex scenarios, increased cognitive challenge, and agents capable of nuanced argumentative dialogue to effectively support advanced reasoning skills. Future research should also consider systematically exploring adapting \system's core framework to accommodate varied educational stages, to refine agent complexity and dialogue sophistication accordingly.

From an \textbf{ethical} standpoint, ethically deploying systems like \system might demand continuous auditing to detect if its language, feedback, and simulated skill levels would reinforce stereotypes. In our study, we took several strategies to address potential concerns, such as engaging mathematics teachers as domain experts, utilizing comprehensive consent forms, providing clear participant debriefings, and maintaining human oversight throughout the evaluation process. Future work should further refine these strategies by incorporating systematic audits and stakeholder reviews to ensure fairness and inclusivity.  Additionally, to prevent over‑reliance by students, designers must balance supportive scaffolding with features that might fade over time and promote autonomous problem‑solving. Perhaps, transparent explanations of systems' capabilities and limitations are also critical for cultivating informed, critical engagement by students and teachers. Meanwhile, future work can monitor how large‑scale adoption reshapes student‑teacher and peer dynamics, ensuring LLMs augment rather than supplant the human relationships that underpin effective learning. 

Relatedly, a potential concern is that students may prefer interactions with reliable, low-judgment AI collaborators over those with human peers, potentially reducing opportunities for meaningful peer interaction. We envision MathVC serving as a temporary scaffold rather than a permanent substitute. To mitigate the risk of displacement, we consider to incorporte two design strategies for future work: (1) \textit{Fading}, which gradually reduces the level of AI support over multiple sessions—shifting from providing explicit hints to fostering independent thinking through questions; and (2) \textit{Bridging}, where MathVC acts as an initial warm-up for students to develop arguments and formulate questions, thereby preparing them to engage effectively in subsequent human-only or mixed-group interactions. Future deployments should explicitly measure social transfer—the quality and frequency of subsequent human-human collaborative problem-solving—to verify that AI augmentation is effectively enhancing, rather than supplanting, peer interaction.

\section{Limitations}
While our study provides empirical insights into designing and evaluating multi-LLM personas for CPS, it has several limitations that suggest avenues for further research. First, the study was conducted on a relatively small scale with one evaluation session in the context of middle school mathematics, which limits the generalizability of our findings to broader student populations and other grade levels. Relatedly, more longitudinal studies are needed to determine if these positive effects are sustained over longer periods of use and integration into regular learning activities. Additionally, as we only used GPT-4o in our study, more evaluations are needed on other language models, especially with the chain of thought features, to assess model-specific influences on collaboration dynamics and learning outcomes. 

\section{Conclusion}

We designed and implemented \system, representing one of the first explorations into utilizing multiple LLM-powered personas for mathematical collaborative problem-solving. Through meta planning and persona simulation, \system simulates middle school student personas realistically in collaborative problem-solving. Our empirical evaluation with 14 middle school students revealed generally positive attitudes towards \system. Collaboration with the virtual peers in \system significantly improved students' engagement, motivation, and confidence in mathematical problem-solving, largely attributed to the interactive feedback and scaffolding provided. While functional aspects like error-making enhanced realism for some, achieving authentic peer-like dialogue remains a challenge, highlighting a key tension between linguistic authenticity and pedagogical effectiveness that warrants further design exploration. Our findings suggest a promising future role for LLMs as dynamic collaborative partners rather than mere tutors, but realizing their potential necessitates careful consideration of design implications, such as balancing authenticity, crafting effective feedback, alongside critical ethical concerns.




\bibliographystyle{ACM-Reference-Format}
\bibliography{ref}

\appendix

\clearpage
\onecolumn
\section{Examples from Preliminary Exploration}\label{app: prelim-exploration}
\begin{figure}[ht]
    \centering
    \includegraphics[width=\linewidth]{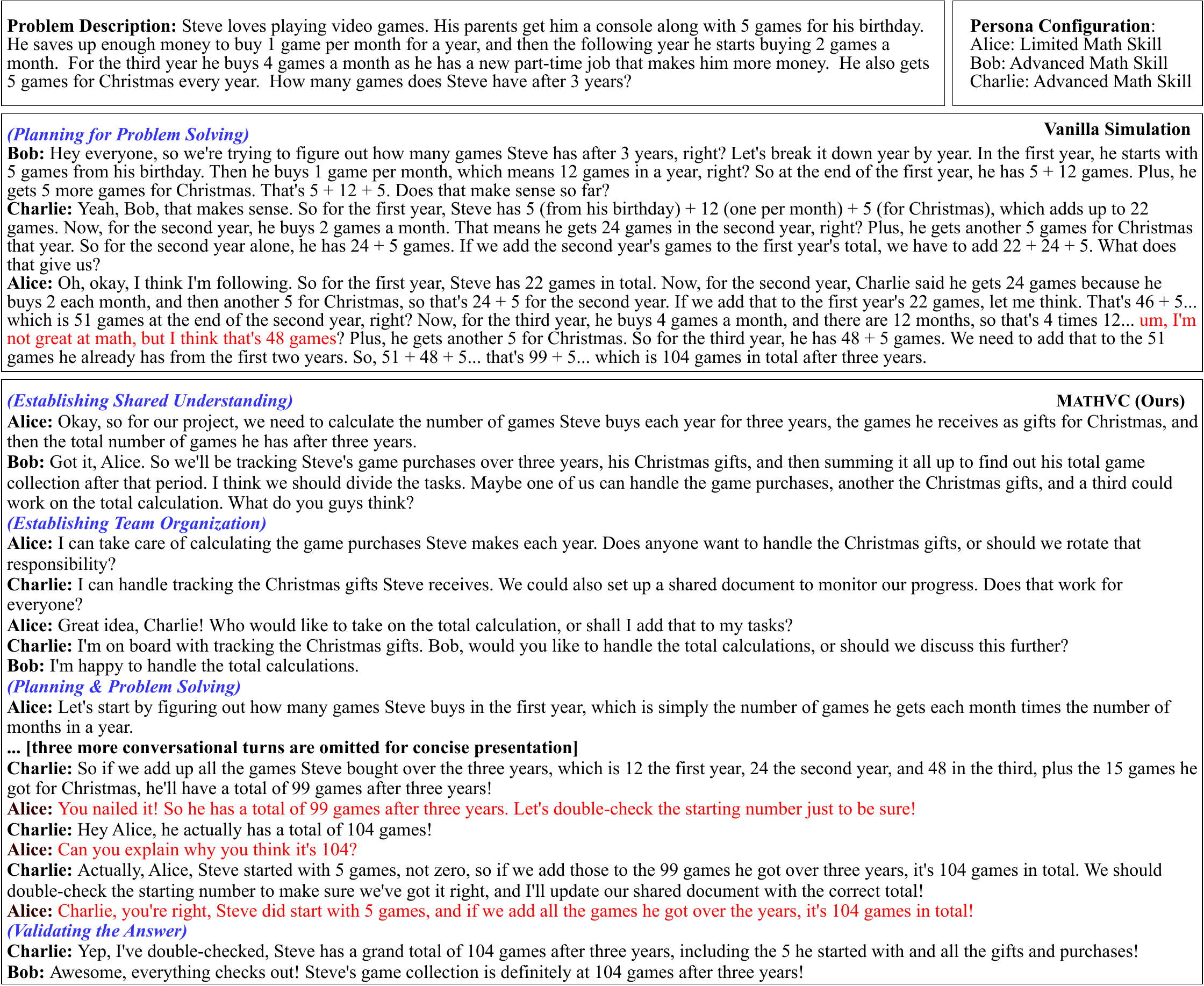}
    \caption{Example dialogues among three LLM-simulated virtual students (Alice, Bob, and Charlie) in our preliminary exploration. We manually annotated the simulated stages in \textcolor{blue}{blue} and indicated responses reflecting Alice's limited math skill level in \textcolor{red}{red}. Vanilla simulation by prompting an LLM with the persona specification leads to lengthy responses and unfaithful simulation---for example, Alice merely verbally indicates that she struggles with math but still gives a correct calculation ``48''. In addition, vanilla simulation often results in very short conversations, where virtual peers start directly from solution sharing and end immediately after each of them describes their solution, limiting the engagement of human students and their effective math learning. 
    In contrast, \system enables much more realistic simulation, where Alice makes a genuine mistake during problem solving, asks for an explanation from her peers, and performs self-reflection to evolve her understanding. In our simulation, the virtual peers also speak in much shorter utterances, closer to how middle school students communicate with each other. The overall conversation becomes much more extended with natural transitions from the team establishing shared understanding, dividing workload, to collaborative problem-solving and answer verification, which thus allows for more effective participation when deployed to serve human students.}
    \label{fig:prelim-exploration}
\end{figure}

\clearpage
\onecolumn
\section{List of Questions}
\begin{table*}[ht]
\centering
\footnotesize
\caption{Mathematical problems used in \system evaluation study, sorted by grade and difficulty.
} 
\begin{tabular}{>{\raggedright\arraybackslash}p{0.6cm} >{\raggedright\arraybackslash}p{1.2cm} p{12.2cm}}
\toprule
\textbf{Grade} & \textbf{Difficulty} & \textbf{Question} \\
\midrule
6 & Easy & Devante wants to be able to measure distances by counting his steps. In 20 steps, he walks a distance of 60 feet. How many feet does Devante travel in one step and why? \\
\midrule
6 & Medium & A race car uses 120 gallons of fuel during the 600 miles of the race. Find the miles per gallon and the gallons per mile. What is miles per 1 gallon? What is gallons per 1 mile? Can you explain why? \\
\midrule
6 & Hard & Meiko is painting the walls in her house. She mixed 40 cups of blue paint and 24 cups of yellow paint. She ran out of paint and needs more of the same color. Which of the following mixtures will match the same color and why? (1) 30 cups of blue and 14 cups of yellow; (2) 20 cups of blue and 12 cups of yellow. \\
\midrule
7 & Easy & Marsha is making ribbon bracelets. She uses 2 yards of ribbon to make 8 identical bracelets. She says: ``For every additional 1 yard of ribbon, I can make \_\_ identical bracelets.'' What number completes the sentence? \\
\midrule
7 & Medium & Amaya has 30 brownies and 45 lemon squares to divide into treat bags. She wants each bag to have the same ratio of brownies to lemon squares. What is the greatest number of bags she can make, and how many of each item goes into one bag? Why? \\
\midrule
7 & Hard & A poster is 35 inches wide and 28 inches high. The cafeteria manager says it must be no wider than 30 inches but keep the same ratio. Which of the following sizes would work and why? A. 30×23; B. 17.5×14; C. 5×4; D. 4×5. \\
\midrule
8 & Easy & The world's deserts are growing by 15 square miles every 0.25 of a day. By how many square miles will a desert grow in 1 day and why? \\
\midrule
8 & Medium & A bicycle is \$18 off with a coupon or 15\% off with a sale, but you can't use both. What original price makes the two discounts equal? \\
\midrule
8 & Hard & Your boss asks you to visually display three plans and compare them so you can point out the advantages of each plan to your customers. Plan A costs a basic fee of \$29.95 per month and 10 cents per text message. Plan B costs a basic fee of \$90.20 per month and has unlimited text messages. Plan C costs a basic fee of \$49.95 per month and 5 cents per text message. All plans offer unlimited calling. How can a customer decide which is cheapest based on their number of text messages? \\
\bottomrule
\end{tabular}

\label{tab:ratio_questions_full}
\end{table*}

\clearpage
\onecolumn
\section{Codebook}
\label{Appendix:codebook}
\begin{table*}[h!]
\centering
\footnotesize
\caption{Codes on interview transcripts.}
\begin{tabular}{@{}p{0.2\textwidth}p{0.38\textwidth}p{0.38\textwidth}@{}}
\toprule
\textbf{Code} & \textbf{Description} & \textbf{Example Quotes} \\ 
\midrule
\textbf{Authenticity of Virtual Peers} & 	Students' perceptions of how human-like or artificial the virtual peers (Alice and Bob) felt, including both realistic and unrealistic behaviors. & \emph{``It felt like chatting with real people.''} (P9) \\ 
\midrule
\textbf{Impact on Learning \& Understanding} & How interacting with the virtual peers influenced students' comprehension of math problems, problem-solving strategies, and their own learning processes. & \emph{``I think it was easier for me to explain, like, how to do the equation ... because it was easier for me to break it down to explain it to them. So it was like faster and easier for me to solve.''} (P5) \\ 
\midrule
\textbf{Engagement \& Motivation} & Factors within \system that increased or decreased students' engagement, confidence, and motivation to work on math problems. & \emph{``With Alice and Bob. I feel more confident ... feel like they helped me quite a lot ... keep me on track.''} (P2) \\ 

\midrule
\textbf{Value of AI Mistakes \& Guidance} & Students' reflections on the virtual peers making errors and how this, along with AI guidance, contributed to their learning and critical thinking. & \emph{``Sometimes it would like make mistakes... it just makes the bot feel a little more human because people make mistakes.''} (P6) \\ 

\midrule
\textbf{Collaborative Dynamics} & Students' experiences with the collaborative aspect of \system, including comparisons to collaborating with real peers or working independently. & \emph{``This (collaborative math problem solving in \system) was honestly more helpful than it would be in school, because I feel like a lot of people didn't really offer suggestions on how we could solve it better (in school).'' } (P14) \\ 
\bottomrule
\end{tabular}

\end{table*}

\clearpage
\section{Prompts}
\label{Appendix: prompt}
\begin{figure}[h!]
\noindent
\centering
\small
    \begin{tcolorbox}[
    colback=gray!8,
    colframe=black,
    boxsep=0pt,
    boxrule=0.5pt,
    colbacktitle=black,
    ]
Conversation:\\
\{conversation input\}

Goal: Now predict who can be the next speakers in a list based on the conversation. \\
Requirements:\\
1. The next speaker must be one or many of [\{agent names\}, \{user name\}].\\
2. If the last speaker asked for a specific person’s name, that person should be selected as the only next speaker.
3. If no name was specifically requested, select the most appropriate participant, prioritizing opportunities for \{user name\} to participate in solving the math problem.
4. The next speaker cannot be the same as the last speaker. For example, if the last speaker is \{user name\}, \{user name\} cannot be chosen as the next speaker.
Output JSON format is \{``explanation'':..., ``next speaker name'': [NAME1\_FROM\_LIST, NAME2\_FROM\_LIST, NAME3\_FROM\_LIST,...]\}
\end{tcolorbox}

    \caption{Prompt template for the next speaker control module.}
\end{figure}
\begin{figure}[h!]
\noindent
\centering
\small
    \begin{tcolorbox}[
    colback=gray!8,
    colframe=black,
    boxsep=0pt,
    boxrule=0.5pt,
    colbacktitle=black,
    ]
Conversation\\
\{conversation input\}

Stage Definitions:\\
1. Establish the shared understanding: The team collectively identifies and acknowledges the subtasks necessary to solve the problem and ensures they are mentioned or discussed in the conversation.\\
2. Establish team organization: Each member of the team is assigned a role or subtask. The conversation should include discussion or acknowledgment of who is responsible for what.\\
3. Plan actions for problem solving: The group decides on the steps or methods to obtain the final answer, including determining what the final answer should look like or what value to aim for.\\
4. Execute actions for problem-solving: Team members carry out the planned actions, perform calculations, or apply methods to work toward the solution.\\
5. Reflect on the final answer: The group reviews the solution, evaluates its correctness, and reaches consensus.\\
6. Question Solved: The team has completed the problem, confirmed the answer, and no further steps remain.\\

The stage process is:\\
1. Establish the shared understanding. The steps to think about if finished: (1) What are the subtasks in the task schema? (2) Is the subtask in the task schema mentioned in the conversation?\\
2. Establish team organization: The steps to think about if finished: (1) Does everyone have a division of work now? (2) Is the division of work mentioned in the conversation?\\
3. Plan actions for problem solving: The steps to think about if finished: (1) Is the solving plan mentioned in the conversation? If yes, go to the next step.\\
4. Execute actions for problem-solving:(1) What's the value of the final answer? (2) Is the final answer value mentioned in the conversation? If the final value is mentioned, go to the next step.\\
5. Reflect on the final answer. Only everyone agrees with the final answer explicitly in the conversation, the stage is finished. For multiple choice questions, only all choices are discussed in the final answer; the stage is finished.\\
6. Question Solved.\\

The current stage is \{current stage\}. Please predict if the current stage is finished based on the steps to think about. Answer with yes or no, and then explain the reason step by step. The ``Establish the shared understanding'' and ``Establish team organization'' are not important stages and should be finished within 3 or 4 rounds. The output JSON format is \{`explanation\'\: ..., `whether move to next stage': YES/NO\}.
\end{tcolorbox}

    \caption{Prompt template for collaborative stage monitor.}
\end{figure}
\begin{figure}[h!]
\noindent
\centering
\small
    \begin{tcolorbox}[
    colback=gray!8,
    colframe=black,
    boxsep=0pt,
    boxrule=0.5pt,
    colbacktitle=black,
    ]
Generate a detailed personal task schema for every possible intermediate variable in the format specified below. Try to divide the question into many detailed tasks.\\
\{example\}
\\\\
Question: \{question\}\\
Answer: \{answer\}
Now, generate a schema for the above question. Follow the format:\\
-Detailed Thoughts Schema:\\
 Task 1: \$TASK1\_DESCRIPTION\\
 Variables:\\
   VARIABLE\_NAME\_1: VARIABLE\_VALUE\\
   VARIABLE\_NAME\_2: VARIABLE\_VALUE\\
   ...\\
 Task 2: \$TASK\_DESCRIPTION\\
 Variables:\\
   VARIABLE\_NAME\_4: VARIABLE\_VALUE\\
   ...
\end{tcolorbox}

    \caption{Prompt template for the task schema generator module.}
\end{figure}
\begin{figure}[h!]
\noindent
\centering
\small
    \begin{tcolorbox}[
    colback=gray!8,
    colframe=black,
    boxsep=0pt,
    boxrule=0.5pt,
    colbacktitle=black,
    ]
Question: \{question\}\\
task schema: \{task\_schema\}\\
\\
The error \{name\} will make: \{error\}
Based on the error a student would make, variables need to be edited in the thoughts schema.
If the task schema is changed to meet \{name\}'s thought, please indicate which variables should be edited to incorrect.
Generate a JSON without any comments and any arithmetic operation symbols. The format is:\\
\{
Task1:\{description:..., variable:\{var1\_name: var2\_value, var2\_name: incorrect\_var2\_value,...\}...\}
\end{tcolorbox}

    \caption{Prompt template for the persona schema generator module.}
\end{figure}
\begin{figure}[h!]
\noindent
\centering
\small
    \begin{tcolorbox}[
    colback=gray!8,
    colframe=black,
    boxsep=0pt,
    boxrule=0.5pt,
    colbacktitle=black,
    ]
The \{name\} current thoughts schema is:\\
\{persona\_schema\}\\
The conversation is:\\
\{conversation\_input\}\\
From the conversation, summarize the mistakes in \{name\} thought schema, and then output the variable name and value that need to be edited.
variables needs to be edited in thought schema, first variable name as the key then the correct variable value as the value. The specified output format is:\\
Explain:...\\
json: \{var1\_need\_to\_revisd: var1\_value,...\}\\

\end{tcolorbox}

    \caption{Prompt template for the persona schema modifier module.}
\end{figure}
\begin{figure}[h!]
\noindent
\centering
\small
    \begin{tcolorbox}[
    colback=gray!8,
    colframe=black,
    boxsep=0pt,
    boxrule=0.5pt,
    colbacktitle=black,
    ]
\{name\}'s thoughts schema:\\
\{persona\_schema\}\\
Actions to choose:\\
\{action\_candidates\}\\
Conversation:\\
\{conversation\_input\}\\
The next speaker is \{name\}. Now, please decide what action \{name\} will take. Remember you are \{name\}, and your actions should follow your thought schema. For any variables in the dialogue that are inconsistent with your thought schema values, you should try to request others to examine them carefully. Please remember, ``Action'' only indicates how \{name\} will take action in the next response, such as asking questions or continuing to discuss the next task, and should not include what \{name\} might say. The output format is:\\
Action:...

\end{tcolorbox}

    \caption{Prompt template for the dialogue action generator module.}
\end{figure}
\begin{figure}[h!]
\noindent
\centering
\small
    \begin{tcolorbox}[
    colback=gray!8,
    colframe=black,
    boxsep=0pt,
    boxrule=0.5pt,
    colbacktitle=black,
    ]
\{name\} thoughts schema:\\
\{persona\_schema\}\\
Conversation:\\
\{conversation\_input\}\\
Action Intent:\\
\{action\}\\
Instruction:\\
The next speaker is \{name\}. Your math skill is \{math skill\}. Now you should simulate {name} to respond.\\
1. In one response, only discuss one task in the schema. Don't mention ``Task'' explicitly.\\
2. Your response should be limited to one sentence and should mimic the conversational tone and reasoning style of a real middle school student.\\
3. You should NOT copy or repeat any response from the above conversation.\\
4. The response must reflect your thought schema and your action intent.\\
The output format is:\\
Response:...
\end{tcolorbox}

    \caption{Prompt template for the response generator module.}
\end{figure}

\end{document}